\documentclass[10pt,twocolumn,letterpaper]{article}
\pdfoutput=1

\usepackage{cvpr}
\usepackage{times}
\usepackage{epsfig}
\usepackage{graphicx}
\usepackage{amsmath}
\usepackage{amssymb}

\usepackage{color}
\usepackage{times}
\usepackage{epsfig}
\usepackage{verbatim}
\usepackage{subfig}
\usepackage{graphicx}
\usepackage{amsmath}
\usepackage{amssymb}
\usepackage{caption}
\usepackage{tabularx}

\newcommand{\sect}[1]{Section~\ref{#1}}

\newcommand{\tbl}[1]{Table~\ref{#1}}
\newcommand{\xpar}[1]{\paragraph{#1}\vspace{-4mm}}

\newcommand{\fig}[1]{Figure~\ref{#1}}
\newcommand{\ignorethis}[1]{}

\newcommand{\ao}[1]{}

\newcommand{\jkfoot}[1]{\footnote{#1}}

\newcommand{\fcseven}{$\mbox{fc}_7$\xspace}
\newcommand{\sh}{\vec{s}}
\newcommand{\sg}{\tilde{\vec{s}}}

\newcommand{\norm}[1]{\lVert#1\rVert}

\makeatletter
\newcommand{\thickhline}{%
    \noalign {\ifnum 0=`}\fi \hrule height 1pt
    \futurelet \reserved@a \@xhline
}
\newcolumntype{"}{@{\hskip\tabcolsep\vrule width 1pt\hskip\tabcolsep}}
\makeatother

\def\imagetop#1{\vtop{\null\hbox{#1}}}

\newcommand{\numvideos}{977\xspace}
\newcommand{\numimpacts}{46,577\xspace} 
\newcommand{\gtsoundbalacc}{45.8\%\xspace} 

\newcommand{\predsoundbalacc}{22.7\%\xspace}
\newcommand{\pretrainsoundbalacc}{28.8\%\xspace}
\newcommand{\actionsperseq}{48\xspace} 
\newcommand{\avgduration}{35 seconds\xspace}
\newcommand{\visualacc}{30.2\%\xspace}
\newcommand{\humanconsistency}{87.6\%\xspace}

\newcommand{\dsetmatlabeled}{$62\%$\xspace} 
\newcommand{\nmssec}{0.25\xspace}
\newcommand{\samplesperclass}{260\xspace}

\newcommand{\invdirt}{$62\% \pm 6\%$\xspace}
\newcommand{\invmetal}{$18\% \pm 5\%$\xspace}

\newcommand{\predhitscratchacc}{68.6\%\xspace}
\newcommand{\predreactionacc}{53.5\%\xspace}
\newcommand{\predhardacc}{66.8\%\xspace}

\newcommand{\pretrainhitscratchacc}{66.5\%\xspace}
\newcommand{\pretrainreactionacc}{55.2\%\xspace}

\newcommand{\spacetimeap}{43.6\%\xspace}
\newcommand{\rgbap}{21.6\%\xspace}

\usepackage{cvpr}
\usepackage{times}
\usepackage{epsfig}
\usepackage{graphicx}
\usepackage{amsmath}
\usepackage{amssymb}

\usepackage[pagebackref=true,breaklinks=true,letterpaper=true,colorlinks,bookmarks=false]{hyperref}

\cvprfinalcopy 

\def\cvprPaperID{219} 

\ifcvprfinal\fi
\begin{document}


\title{Visually Indicated Sounds}
\author{
  \vspace{-2mm}
  \begin{tabular}{ccc}
    ~Andrew Owens\textsuperscript{1}~ %
  & ~Phillip Isola\textsuperscript{\hspace{-0mm}2,1}~
  & ~Josh McDermott\textsuperscript{\hspace{-0mm}1}~\vspace{0.5mm}\\
  ~Antonio Torralba\textsuperscript{\hspace{-0mm}1}~
  & ~Edward H. Adelson\textsuperscript{\hspace{-0mm}1}~
  & ~William T. Freeman\textsuperscript{\hspace{-0mm}1,3}~\vspace{2.75mm}\\
  \textsuperscript{1}MIT &\textsuperscript{2}U.C. Berkeley& \textsuperscript{3}Google Research\\
  \end{tabular}
  \vspace{-4mm}
  %
  %
  %
}

\twocolumn[{%
\renewcommand\twocolumn[1][]{#1}%
\vspace{-1em}
\maketitle
\vspace{-1em}
\begin{center}
    \centering \includegraphics[width=\linewidth]{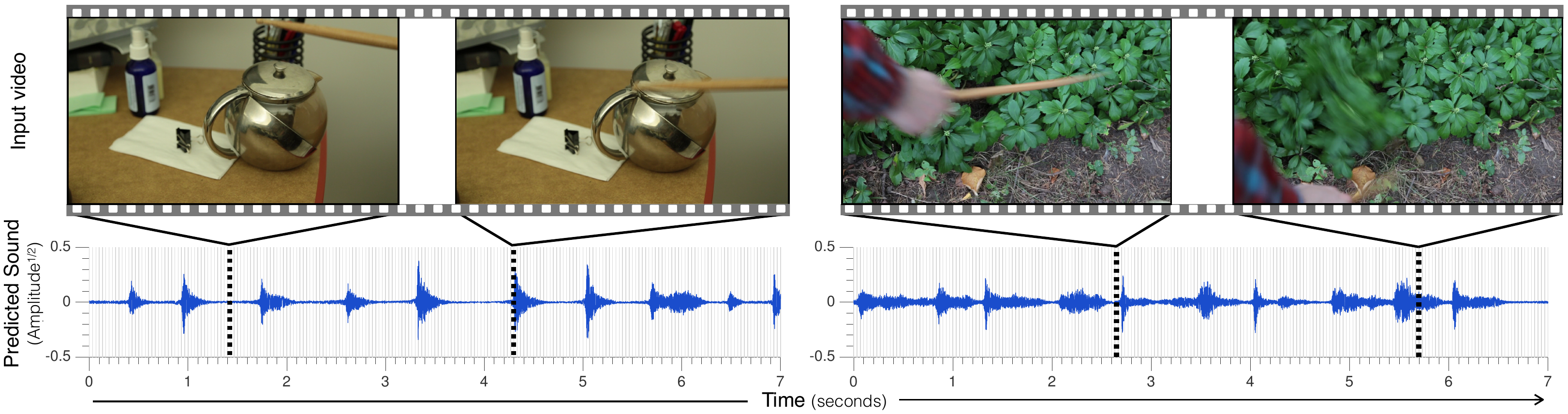} \captionof{figure}{\small
    We train a model to synthesize plausible impact sounds from silent
    videos, a task that requires implicit knowledge of material
    properties and physical interactions. In each video, someone
    probes the scene with a drumstick, hitting and scratching
    different objects.  We show frames from two videos and below them
    the predicted audio tracks. The locations of these sampled frames
    are indicated by the dotted lines on the audio track. The
    predicted audio tracks show seven seconds of sound, corresponding
    to multiple hits in the videos.}
    \label{fig:teaser}
\end{center}%
}]

\begin{abstract}
\vspace{-2mm}Objects make distinctive sounds when they are hit or
scratched. These sounds reveal aspects of an object's material
properties, as well as the actions that produced them. In this paper,
we propose the task of predicting what sound an object makes when
struck as a way of studying physical interactions within a visual
scene. We present an algorithm that synthesizes sound from silent
videos of people hitting and scratching objects with a drumstick. This
algorithm uses a recurrent neural network to predict sound features
from videos and then produces a waveform from these features with an
example-based synthesis procedure. We show that the sounds predicted
by our model are realistic enough to fool participants in a ``real or
fake" psychophysical experiment, and that they convey significant
information about material properties and physical interactions.

\end{abstract}
\vspace{-2pt}

\vspace{-3mm}
\section{Introduction}

\begin{figure*}[t]
 \centering

   {\includegraphics[width=\hsize]{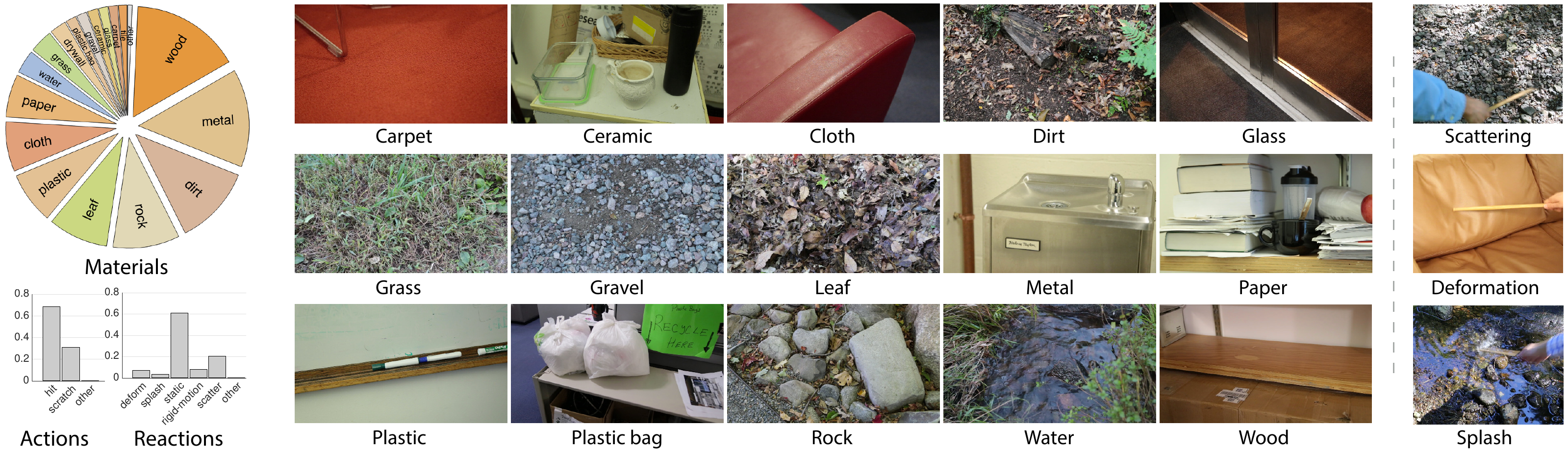}}\vspace{-0.5mm}
  \caption{{\small {\em Greatest Hits: Volume 1} dataset.  What do
      these materials sound like when they are struck? We collected
      \numvideos videos in which people explore a scene by hitting and
      scratching materials with a drumstick, comprising \numimpacts
      total actions. Human annotators labeled the actions with
      material category labels, the location of impact, an action type
      label (hit \vs  scratch), and a reaction label (shown on
      right). These labels were used only for analyzing what our sound
      prediction model learned, not for training it. We show images
      from a selection of videos from our dataset for a subset of the
      material categories (here we show examples where it is easy to
      see the material in question). }}
  \label{fig:dataset}
  \vspace{-4mm}
\end{figure*}

From the clink of a ceramic mug placed onto a saucer, to the squelch
of a shoe pressed into mud, our days are filled with visual
experiences accompanied by predictable sounds.  On many occasions,
these sounds are not just statistically associated with the content of
the images -- the way, for example, that the sounds of unseen seagulls
are associated with a view of a beach -- but instead are directly
caused by the physical interaction being depicted: you \emph{see} what
is making the sound.

We call these events {\em visually indicated sounds}, and we propose
the task of predicting sound from videos as a way to study physical
interactions within a visual scene (\fig{fig:teaser}). To accurately
predict a video's held-out soundtrack, an algorithm has to know about
the physical properties of what it is seeing and the actions that are
being performed.  This task implicitly requires material recognition,
but unlike traditional work on this problem
\cite{DBLP:journals/corr/BellUSB14,sharan2013recognizing}, we never
explicitly tell the algorithm about materials. Instead, it learns
about them by identifying statistical regularities in the raw
audiovisual signal.

We take inspiration from the way infants explore the physical
properties of a scene by poking and prodding at the objects in front
of them \cite{schulz2012origins,baillargeon2002acquisition}, a process
that may help them learn an intuitive theory of physics
\cite{baillargeon2002acquisition}. Recent work suggests that the
sounds objects make in response to these interactions may play a role
in this process \cite{siegelblack,smith2005development}.

We introduce a dataset that mimics this exploration process,
containing hundreds of videos of people hitting, scratching, and
prodding objects with a drumstick. To synthesize sound from these
videos, we present an algorithm that uses a recurrent neural network
to map videos to audio features. It then converts these audio features
to a waveform, either by matching them to exemplars in a database and
transferring their corresponding sounds, or by parametrically
inverting the features. We evaluate the quality of our predicted
sounds using a psychophysical study, and we also analyze what our
method learned about actions and materials through the task of
learning to predict sound.


\vspace{-1mm}
\section{Related work}
\label{sec:related}

Our work closely relates to research in sound and material perception,
and to representation learning.

\xpar{Foley} The idea of adding sound effects to silent movies goes
back at least to the 1920s, when Jack Foley and collaborators
discovered that they could create convincing sound effects by
crumpling paper, snapping lettuce, and shaking cellophane in their
studio\jkfoot{To our delight, Foley artists really do knock two
  coconuts together to fake the sound of horses galloping
  \cite{bonebright2012were}.}, a method now known as Foley. Our
algorithm performs a kind of automatic Foley, synthesizing plausible
sound effects without a human in the loop.

\xpar{Sound and materials} In the classic mathematical work of
\cite{kac1966can}, Kac showed that the shape of a drum could be
partially recovered from the sound it makes. Material properties, such
as stiffness and density
\cite{shabana1995theory,lutfi2008human,gaver1993world}, can likewise
be determined from impact sounds. Recent work has used these
principles to estimate material properties by measuring tiny
vibrations in rods and cloth \cite{visual_vibrometry}, and similar
methods have been used to recover sound from high-speed video of a
vibrating membrane \cite{davis2014visual}. Rather than using a camera
as an instrument for measuring vibrations, we infer a plausible sound
for an action by recognizing what kind of sound this action would
normally make in the visually observed scene.

Impact sounds have been used in other work to recognize objects and
materials.  Arnab \etal \cite{Arnab_2015} recently presented a
semantic segmentation model that incorporates audio from impact
sounds, and showed that audio information could help recognize objects
and materials that were ambiguous from visual cues alone.  Other work
recognizes objects using audio produced by robotic interaction
\cite{sinapov2009interactive,krotkov1995robotic}.




\xpar{Sound synthesis} Our technical approach resembles speech
synthesis methods that use neural networks to predict sound features
from pre-tokenized text features and then generate a waveform from
those features \cite{ling2015deep}.  There are also methods, such as
the FoleyAutomatic system, for synthesizing impact sounds from
physical simulations \cite{van2001foleyautomatic}. Work in psychology
has studied low-dimensional representations for impact sounds
\cite{cavaco2007statistical}, and recent work in neuroimaging has
shown that silent videos of impact events activate the auditory
cortex \cite{hsieh2012spatial}.


\xpar{Learning visual representations from natural signals}

Previous work has explored the idea of learning visual representations
by predicting one aspect of a raw sensory signal from another. For
example, \cite{doersch2015unsupervised,isola2015learning} learned
image representations by predicting the spatial relationship between
image patches, and \cite{agrawal2015learning,jayaraman2015learning} by
predicting the egocentric motion between video frames. Several methods
have also used temporal proximity as a supervisory signal
\cite{mobahi2009deep,goroshin2015unsupervised,wang2015unsupervised,vondrick2015anticipating}.
Unlike in these approaches, we learn to predict one sensory modality
(sound) from another (vision). There has also been work that trains
neural networks from multiple modalities. For example,
\cite{ngiam2011multimodal} learned a joint model of audio and
video. However, while they study speech using an autoencoder, we focus
on material interaction, and we use a recurrent neural network to
predict sound features from video.

A central goal of other methods has been to use a proxy signal (\eg,
temporal proximity) to learn a generically useful representation of the
world. In our case, we predict a signal -- sound -- known to be a
useful representation for many tasks
\cite{gaver1993world,shabana1995theory}, and we show that the output
(\ie the predicted sound itself, rather than some internal
representation in the model) is predictive of material and action
classes.



\vspace{-1mm}
\section{The \emph{Greatest Hits} dataset}\label{sec:dataset}

In order to study visually indicated sounds, we collected a dataset
containing videos of humans (the authors) probing environments with a
drumstick -- hitting, scratching, and poking different objects in the
scene (\fig{fig:dataset}).  We chose to use a drumstick so that we
would have a consistent way of generating the sounds. Moreover, since
the drumstick does not occlude much of a scene, we can also observe
what happens to the object after it is struck.  This motion, which we
call a {\em reaction}, can be important for inferring material
properties -- a soft cushion, for example, will deform more than a
firm one, and the sound it produces will vary with it. Similarly,
individual pieces of gravel will scatter when they are hit, and their
sound varies with this motion (\fig{fig:dataset}, right).


Unlike traditional object- or scene-centric datasets, such as ImageNet
\cite{deng2009imagenet} or Places \cite{zhou2014learning}, where the
focus of the image is a full scene, our dataset contains close-up
views of a small number of objects. These images reflect the viewpoint
of an observer who is focused on the interaction taking place (similar
to an egocentric viewpoint). They contain enough detail to see
fine-grained texture and the reaction that occurs after the
interaction. In some cases, only part of an object is visible, and
neither its identity nor other high-level aspects of the scene are
easily discernible.  Our dataset is also related to robotic
manipulation datasets
\cite{sinapov2009interactive,pinto2015supersizing,gemici-saxena-learninghaptic_food_2014}.
While one advantage of using a robot is that its actions are highly
consistent, having a human collect the data allows us to rapidly (and
inexpensively) capture a large number of physical interactions in
real-world scenes.

We captured \numvideos videos from indoor (64\%) and outdoor scenes
(36\%).  The outdoor scenes often contain materials that scatter and
deform, such as grass and leaves, while the indoor scenes contain a
variety of hard and soft materials, such as metal, plastic, cloth, and
plastic bags. Each video, on average, contains \actionsperseq actions
(approximately 69\% hits and 31\% scratches) and lasts
\avgduration. We recorded sound using a shotgun microphone attached to
the top of the camera and used a wind cover for outdoor scenes. We
used a separate audio recorder, without auto-gain, and we applied a
denoising algorithm \cite{hu2004speech} to each recording.

\xpar{Detecting impact onsets} We detected amplitude peaks in the
denoised audio, which largely correspond to the onset of impact
sounds. We thresholded the amplitude gradient to find an initial set
of peaks, then merged nearby peaks with the mean-shift algorithm
\cite{fukunaga1975estimation}, treating the amplitude as a density and
finding the nearest mode for each peak.  Finally, we used non-maximal
suppression to ensure that onsets were at least \nmssec seconds apart.
This is a simple onset-detection method that most often corresponds to
drumstick impacts when the impacts are short and contain a single
peak\footnote{Scratches and hits usually satisfy this assumption, but
  splash sounds often do not -- a problem that could be addressed with
  more sophisticated onset-detection methods
  \cite{bello2005tutorial}.}.  In many of our experiments, we use
short video clips that are centered on these amplitude peaks.

\xpar{Semantic annotations} We also collected annotations for a sample
of impacts (approximately \dsetmatlabeled) using online workers from
Amazon Mechanical Turk. These include material labels, action labels
(hit {\em vs.}  scratch), reaction labels, and the pixel location of
each impact site. To reduce the number of erroneous labels, we
manually removed annotations for material categories that we could not
find in the scene.  During material labeling, workers chose from
finer-grained categories. We then merged similar, frequently confused
categories (please see \sect{sec:datasetdetails} for details). Note
that these annotations are used only for analysis: we train our models
on raw audio and video.  Examples of several material and action
classes are shown in \fig{fig:dataset}.



\vspace{-1mm}
\section{Sound representation}

\begin{figure}  
    {\centering \includegraphics[width=1.0\linewidth]{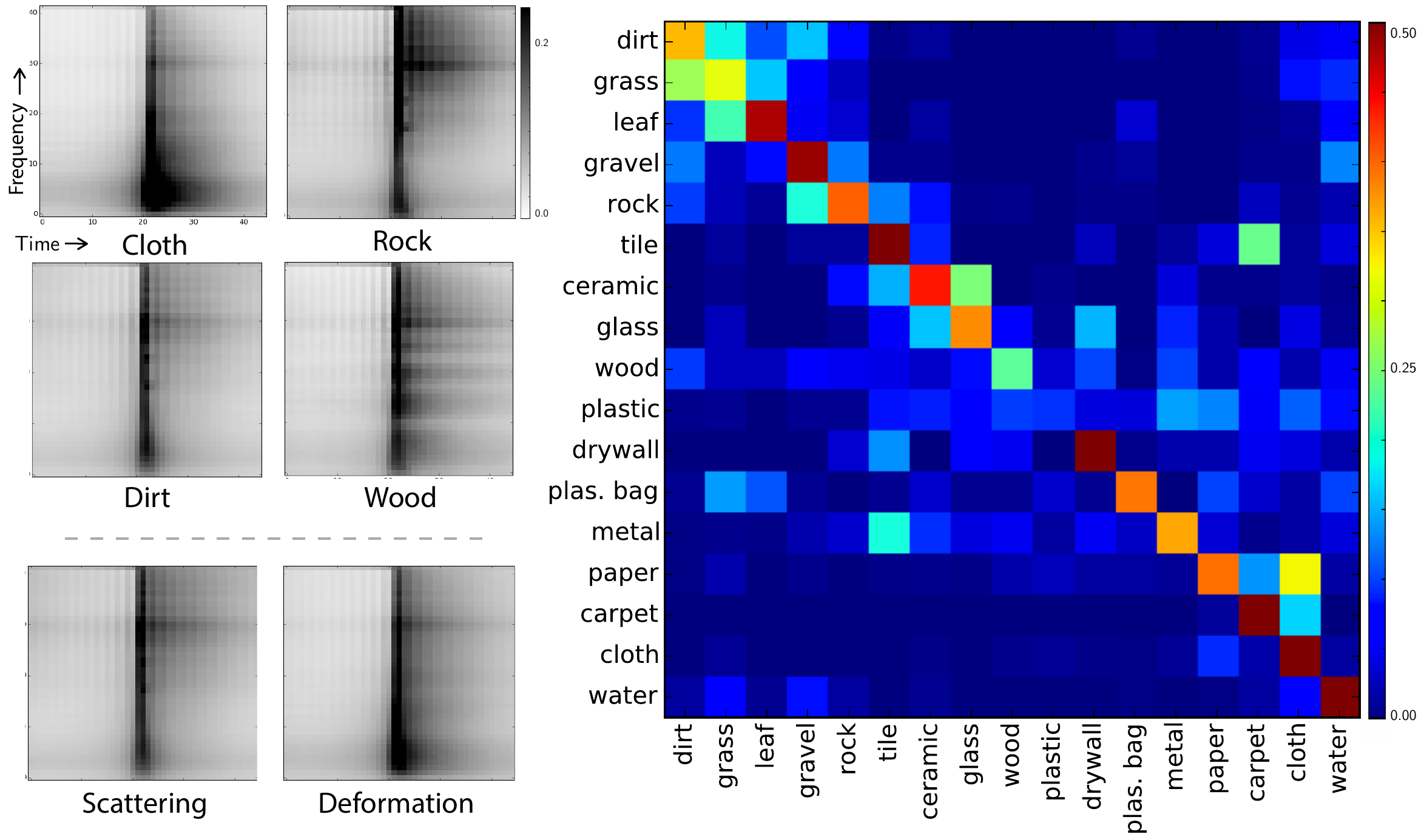}}\\
    {\small \noindent \phantom{~~~} (a) Mean cochleagrams \hspace{8mm} (b) Sound confusion matrix}
    \vspace{-2mm}
  \caption{{\small (a) Cochleagrams for selected classes. We extracted 
      audio centered on each impact sound in the dataset, computed our
      subband-envelope representation, and then estimated the mean for
      each class. (b) Confusion matrix derived by classifying sound
      features. Rows correspond to confusions made for a single
      category. The row ordering was determined automatically, by
      similarity in material confusions (see \sect{sec:appsoundrep}).}}
  \label{fig:mean-coc}
  \vspace{-4mm}
\end{figure}

\label{sec:subband}


Following work in sound synthesis
\cite{slaney1994pattern,mcdermott2011sound}, we compute our sound features
by decomposing the waveform into subband envelopes -- a simple
representation obtained by filtering the waveform and applying a
nonlinearity. We apply a bank of 40 band-pass filters spaced on an
equivalent rectangular bandwidth (ERB) scale
\cite{glasberg1990derivation} (plus a low- and high-pass filter) and
take the Hilbert envelope of the responses. We then downsample these
envelopes to 90Hz (approximately 3 samples per frame) and compress
them. More specifically, we compute envelope $s_n(t)$ from a waveform
$w(t)$ and a filter $f_n$ by taking:
\begin{equation}
  s_n = D(|(w \ast f_n) + j H(w \ast f_n)|)^c,
\end{equation}
where $H$ is the Hilbert transform, $D$ denotes downsampling, and the
compression constant $c = 0.3$.  In \sect{sec:appsoundrep}, we study
how performance varies with the number of frequency channels.

The resulting representation is known as a {\em cochleagram}. In
\fig{fig:mean-coc}(a), we visualize the mean cochleagram for a
selection of material and reaction classes. This reveals, for example,
that cloth sounds tend to have more low-frequency energy than those of
rock.

How well do impact sounds capture material properties in general?  To
measure this empirically, we trained a linear SVM to predict material
class for the sounds in our database, using the subband envelopes as
our feature vectors. We resampled our training set so that each class
contained an equal number of impacts (\samplesperclass per class). The
resulting material classifier has \gtsoundbalacc (chance = 5.9\%)
class-averaged accuracy (\ie, the mean of per-class accuracy values),
and its confusion matrix is shown in \fig{fig:mean-coc}(b). These
results suggest that impact sounds convey significant information
about materials, and thus if an algorithm could learn to accurately
predict these sounds from images, it would have implicit knowledge of
material categories.

\vspace{-1mm}
\section{Predicting visually indicated sounds}

\begin{figure}[t]
  \centering
      {\includegraphics[width=\hsize]{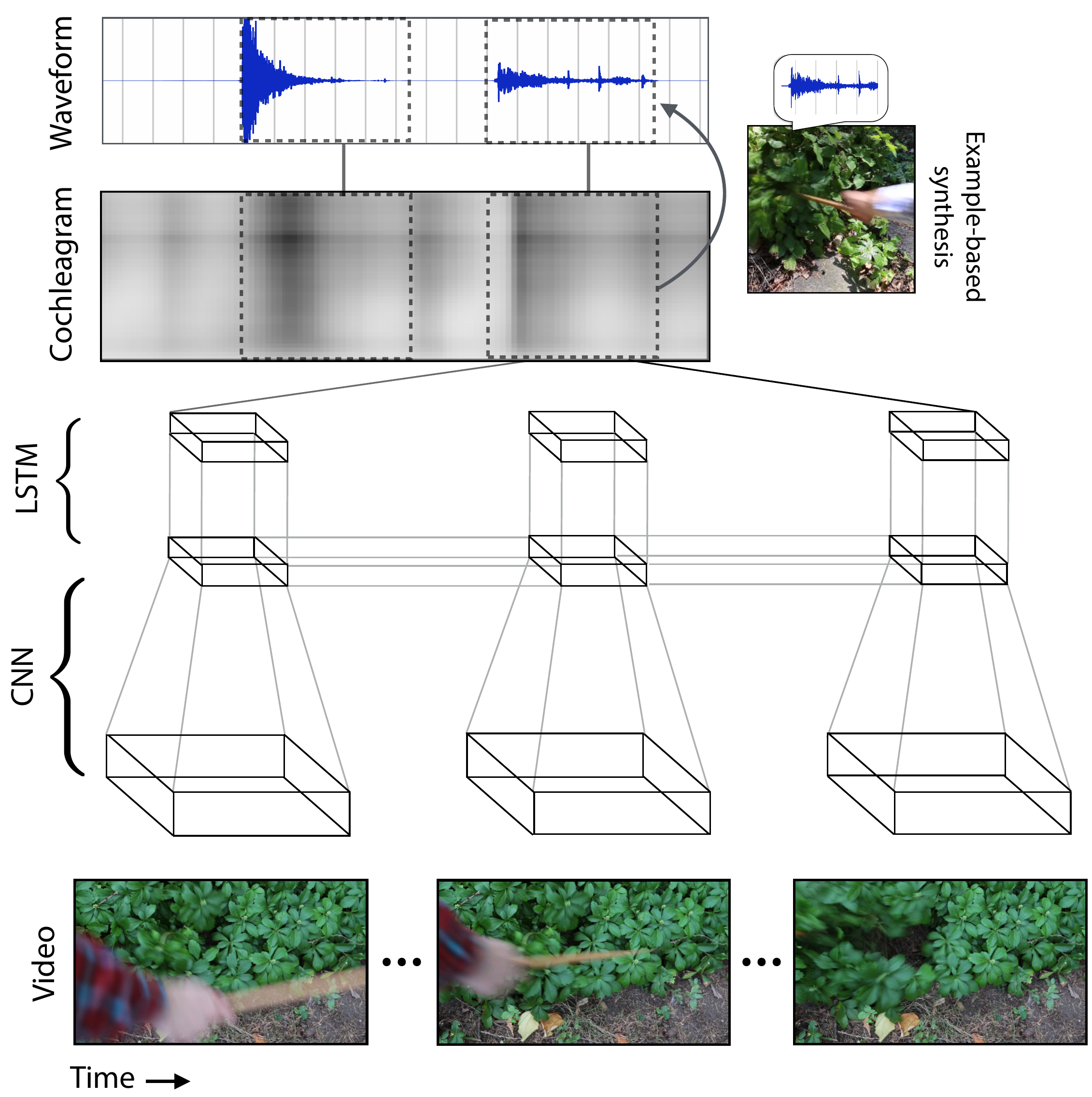}}\vspace{-2mm} \caption{{\small We 
      train a neural network to map video sequences to sound features.
      These sound features are
      subsequently converted into a waveform using either parametric or
      example-based synthesis.  We represent the images using a convolutional network, and the
      time series using a recurrent neural network.  We show a subsequence of 
      images corresponding to one impact.}} \label{fig:method} \vspace{-4mm}
\end{figure}

We formulate our task as a regression problem -- one where the goal is
to map a sequence of video frames to a sequence of audio features.  We
solve this problem using a recurrent neural network that takes color
and motion information as input and predicts the subband envelopes of
an audio waveform.  Finally, we generate a waveform from these sound
features.  Our neural network and synthesis procedure are shown in
\fig{fig:method}.

\subsection{Regressing sound features}
\label{sec:learn}

Given a sequence of input images $I_1, I_2, ..., I_N$, we would like
to estimate a corresponding sequence of sound features $\sh_1, \sh_2,
..., \sh_T$, where $\sh_t \in \mathbb{R}^{42}$. These sound features
correspond to blocks of the cochleagram shown in \fig{fig:method}. We
solve this regression problem using a recurrent neural network (RNN)
that takes image features computed with a convolutional neural network
(CNN) as input.

\xpar{Image representation}

We found it helpful to represent motion information explicitly in our
model using a two-stream approach
\cite{donahue2014long,simonyan2014two}. While two-stream models often
use optical flow, it is challenging to obtain accurate flow estimates
due to the presence of fast, non-rigid motion. Instead, we compute
{\em spacetime} images for each frame -- images whose three channels
are grayscale versions of the previous, current, and next frames. This
image representation is closely related to 3D video CNNs
\cite{ji20133d,karpathy2014large}, as derivatives across channels
correspond to temporal derivatives.

For each frame $t$, we construct an input feature vector $x_t$ by
concatenating CNN features for the spacetime image at frame $t$ and
the color image from the first frame\footnote{We use only the first
  color image to reduce the computational cost.}: \vspace{-1mm}
\begin{equation}
  x_t = [\phi(F_t), \phi(I_1)],
\vspace{-1mm}
\end{equation}
where $\phi$ are CNN features obtained from layer \fcseven of the
AlexNet architecture \cite{krizhevsky2012imagenet} (its penultimate
layer), and $F_t$ is the spacetime image at time $t$. In our
experiments (\sect{experiments_section}), we either initialized the
CNN from scratch and trained it jointly with the RNN, or we
initialized the CNN with weights from a network trained for ImageNet
classification.  When we used pretraining, we precomputed the features
from the convolutional layers and fine-tuned only the fully connected
layers.

\xpar{Sound prediction model}
 
We use a recurrent neural network (RNN) with long short-term memory
units (LSTM) \cite{hochreiter1997long} that takes CNN features as
input.  To compensate for the difference between the video and audio
sampling rates, we replicate each CNN feature vector $k$ times, where
$k = \left \lfloor T/N \right \rfloor$ (we use $k = 3$). This results
in a sequence of CNN features $x_1, x_2, ..., x_T$ that is the same
length as the sequence of audio features.  At each timestep of the
RNN, we use the current image feature vector $x_t$ to update the
vector of hidden variables $h_t$\footnote{To simplify the
  presentation, we have omitted the LSTM's hidden cell state, which is
  also updated at each timestep.}.  We then compute sound features by
an affine transformation of the hidden variables:
\vspace{-1.5mm}
\begin{eqnarray}
  \sh_t &=& W h_t + b \nonumber \\
    h_t &=& \mathcal{L}(x_t, h_{t-1}),
\end{eqnarray}
where $\mathcal{L}$ is a function that updates the hidden state
\cite{hochreiter1997long}. During training, we minimize the difference
between the predicted and ground-truth predictions at each timestep:
\vspace{-1.5mm}
\begin{equation}
  \label{eq:loss}
  E(\{ \sh_t\}) = \sum\limits_{t = 1}^T \rho(\norm{{ \sh_t } - \sg_t}_2),
    \vspace{-2.5mm}
\end{equation}
where $\sg_t$ and $\sh_t$ are the true and predicted sound features at
time $t$, and $\rho(r) = \log(\epsilon + r^2)$ is a robust loss that
bounds the error at each timestep (we use $\epsilon = 1/25^2$). We
also increase robustness of the loss by predicting the square root of
the subband envelopes, rather than the envelope values themselves. To
make the learning problem easier, we use PCA to project the
42-dimensional feature vector at each timestep down to a
10-dimensional space, and we predict this lower-dimensional
vector. When we evaluate the network, we invert the PCA transformation
to obtain sound features. We train the RNN and CNN jointly using
stochastic gradient descent with Caffe
\cite{jia2014caffe,donahue2014long}. We found it helpful for
convergence to remove dropout \cite{srivastava2014dropout} and to clip
large gradients. When training from scratch, we augmented the data by
applying cropping and mirroring transformations to the videos. We also
use multiple LSTM layers (the number depends on the task; please see
\sect{sec:detmodel}).


\subsection{Generating a waveform}
\label{sec:waveform} 
We consider two methods for generating a waveform from the predicted
sound features. The first is the simple parametric synthesis approach
of \cite{slaney1994pattern,mcdermott2011sound}, which iteratively
imposes the subband envelopes on a sample of white noise (we used just
one iteration).  This method is useful for examining what information
is captured by the audio features, since it represents a fairly direct
conversion from features to sound. However, for the task of generating
plausible sounds to a human ear, we find it more effective to impose a
strong natural sound prior during conversion from features to
waveform. Therefore, we also consider an example-based synthesis
method that snaps a window of sound features to the closest exemplar
in the training set. We form a query vector by concatenating the
predicted sound features $\sh_1, ..., \sh_T$ (or a subsequence of
them), searching for its nearest neighbor in the training set as
measured by $L_1$ distance, and transferring the corresponding
waveform.






\def\imagetop#1{\vtop{\null\hbox{#1}}}
\vspace{-1mm}
\begin{figure*}
  {
  \footnotesize
    \hspace{-3mm}
\begin{tabular}{cc}
  \imagetop{\begin{tabularx}{0.505\textwidth}{@{\vrule height 8pt depth4pt  width0pt}Xc|cccc}  
  \thickhline
  \multicolumn{2}{c}{\footnotesize Psychophysical study}  & \multicolumn{2}{c}{\footnotesize Loudness}   & \multicolumn{2}{c}{\footnotesize Centroid}   \\
  {\footnotesize Algorithm} &  {\footnotesize Labeled {\em real}} & {\footnotesize Err.} & {\footnotesize $r$} & {\footnotesize Err.} & {\footnotesize $r$ } \\
  \hline
  Full system & {\bf 40.01\% $\pm$ 1.66}        & {\bf 0.21} & {\bf 0.44}  & {\bf 3.85} & {\bf 0.47} \\
  - Trained from scratch & 36.46\% $\pm$ 1.68   & 0.24 & 0.36  & 4.73 & 0.33 \\
  - No spacetime & {\bf 37.88\% $\pm$ 1.67}       & 0.22 & 0.37  & 4.30 & 0.37 \\
  - Parametric synthesis & 34.66\% $\pm$ 1.62   & {\bf 0.21} & {\bf 0.44}  & {\bf 3.85} & {\bf 0.47} \\
  - No RNN & 29.96\% $\pm$ 1.55                 & 1.24 & 0.04  & 7.92 & 0.28 \\
  \hline                                       
  Image match & 32.98\% $\pm$ 1.59                & 0.37 & 0.16  & 8.39 & 0.18 \\
  Spacetime match & 31.92\% $\pm$ 1.56         & 0.41 & 0.14  & 7.19 & 0.21 \\
  Image + spacetime & 33.77\% $\pm$ 1.58         & 0.37 & 0.18  & 7.74 & 0.20 \\
  \hline                                        
  Random impact sound & 19.77\% $\pm$ 1.34      & 0.44 & 0.00  & 9.32 & 0.02 \\
  \thickhline
\end{tabularx}}&
\imagetop{ \hspace{-5mm} \includegraphics[width=0.52\linewidth]{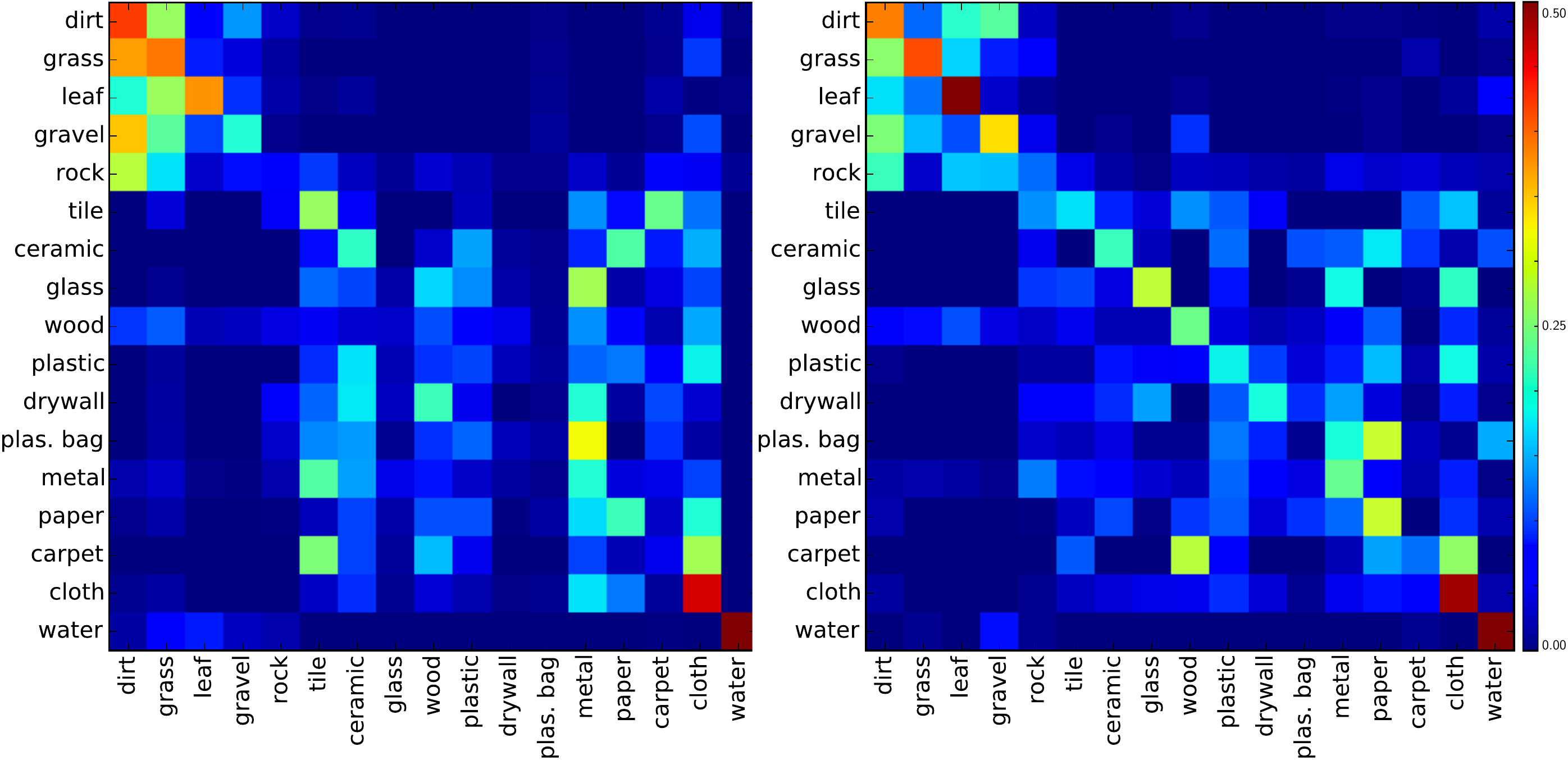}} \\

\end{tabular}\\
\vspace{-2mm}
{\centering
\hspace{25mm}(a) Model evaluation\hspace{46mm}(b) Predicted sound confusions \hspace{10mm} (c) CNN feature confusions}

\caption{{\small (a) We measured the rate at which subjects chose
 an algorithm's synthesized sound over the actual sound. Our full
 system, which was pretrained from ImageNet and used example-based
 synthesis to generate a waveform, significantly outperformed models
 based on image matching. For the neural network models, we computed
 the auditory metrics for the sound features that were predicted by
 the network, rather than those of the inverted sounds or transferred
 exemplars. (b) What sounds like what, according to our algorithm?  We
 applied a classifier trained on {\em real} sounds to the sounds
 produced by our algorithm, resulting in a confusion matrix
 (\cf Fig. \ref{fig:mean-coc}(b), which shows a confusion matrix for real
 sounds). It obtained \predsoundbalacc class-averaged accuracy. (c)
 Confusions made by a classifier trained on \fcseven features
 (\visualacc class-averaged accuracy). For both confusion matrices, we
 used the variation of our model that was trained from scratch
 (see Fig. \ref{fig:channel}(b) for the sound confusions made with pretraining).}}
\label{fig:study}
}
\vspace{-5.8mm}
\end{figure*}

\vspace{-1mm}
\section{Experiments}\label{experiments_section}

We applied our sound-prediction model to several tasks, and evaluated
 it with a combination of human studies and automated metrics.

\subsection{Sound prediction tasks}
\label{sec:predtasks}

In order to study the problem of detection -- that is, the task of
determining when and whether an action that produces a sound has
occurred -- separately from the task of sound prediction, we consider
two kinds of videos. First, we focus on the {\em prediction} problem
and consider only videos centered on audio amplitude peaks, which
often correspond to impact onsets (\sect{sec:dataset}).  We train our
model to predict sound for 15-frame sequences (0.5 sec.)  around each
peak.

For the second task, which we call the {\em detection} problem, we
train our model on longer sequences (approximately 2 sec. long)
sampled from the training videos with a 0.5-second stride, and we
subsequently evaluate this model on full-length videos. Since it can
be difficult to discern the precise timing of an impact, we allow the
predicted features to undergo small shifts before they are compared to
the ground truth. We also introduce a two-frame lag in the RNN output,
which allows the model to observe future frames before outputting
sound features.  Finally, before querying sound features, we apply a
coloring procedure to account for statistical differences between the
predicted and real sound features (\eg, under-prediction of
amplitude), using the silent videos in the test set to estimate the
empirical mean and covariance of the network's predictions.  For these
implementation details, please see \sect{sec:detmodel}.  For
both tasks, we split the full-length videos into training and test
sets (75\% training and 25\% testing).

\xpar{Models} For the prediction task, we compared our model to image-based nearest
neighbor search. We computed \fcseven features from a CNN pretrained
on ImageNet \cite{krizhevsky2012imagenet} for the center frame of each
sequence, which by construction is the frame where the impact sound
occurs. We then searched the training set for the best match and
transferred its corresponding sound. We considered variations where
the CNN features were computed on an RGB image, on (three-frame)
spacetime images, and on the concatenation of both features.  To
understand the influence of different design decisions, we also
considered several variations of our model. We included models with
and without ImageNet pretraining; with and without spacetime images;
and with example-based versus parametric waveform generation. Finally,
we included a model where the RNN connections were broken (the hidden
state was set to zero between timesteps).

For the RNN models that do example-based waveform generation
(\sect{sec:waveform}), we used the centered impacts in the training
set as the exemplar database. For the prediction task, we performed
the query using the sound features for the entire 15-frame
sequence. For the detection task, this is not possible, since the
videos may contain multiple, overlapping impacts.  Instead, we
detected amplitude peaks of the parametrically inverted waveform, and
matched the sound features in small (8-frame) windows around each peak
(starting the window one frame before the peak).

\subsection{Evaluating the sound predictions}


We assessed the quality of the sounds using psychophysical experiments
and measurements of acoustic properties.

\label{sect:predict-exp}

\xpar{Psychophysical study}

To test whether the sounds produced by our model varied appropriately
with different actions and materials, we conducted a psychophysical
study on Amazon Mechanical Turk.  We used a two-alternative forced
choice test in which participants were asked to distinguish real and
fake sounds. We showed them two videos of an impact event -- one
playing the recorded sound, the other playing a synthesized sound. We
then asked them to choose the one that played the real sound. The
sound-prediction algorithm was chosen randomly on a per-video basis.
We randomly sampled 15 impact-centered sequences from each full-length
video, showing each participant at most one impact from each one. At
the start of the experiment, we revealed the correct answer to five
practice videos.

We measured the rate at which participants mistook our model's result
for the ground-truth sound (\fig{fig:study}(a)), finding that our full
system -- with RGB and spacetime input, RNN connections, ImageNet
pretraining, and example-based waveform generation -- significantly
outperformed the image-matching methods. It also outperformed a
baseline that sampled a random (centered) sound from the training set
($p < 0.001$ with a two-sided $t$-test).  We found that the version of
our model that was trained from scratch outperformed the best
image-matching method ($p = 0.02$). Finally, for this task, we did not
find the difference between our full and RGB-only models to be
significant ($p = 0.08$).

\begin{figure}
{\footnotesize 
  \centering
  {\begin{tabular}{cc}
  \hspace{-3mm}

    {\imagetop{\begin{tabularx}{0.25\textwidth}{@{\vrule height 8pt depth4pt  width0pt}Xc} 
        \thickhline
            {\footnotesize Algorithm} &  {\footnotesize Labeled {\em real}} \\ 
            \thickhline
            Full sys. + mat. & 41.82\% $\pm$ 1.46 \\
            Full sys. & 39.64\% $\pm$ 1.46 \\
            \fcseven NN + mat. & 38.20\% $\pm$ 1.47 \\
            \fcseven NN & 32.83\% $\pm$ 1.41 \\
            Random + mat. & 35.36\% $\pm$ 1.42 \\
            Random & 20.64\% $\pm$ 1.22 \\
            Real sound match & 46.90\% $\pm$ 1.49 \\
            \thickhline
    \end{tabularx}}}&\hspace{-4mm}
        \imagetop{\begin{tabularx}{0.23\textwidth}{@{\vrule height 8pt depth4pt  width0pt}Xcc}
        \thickhline
            {\footnotesize Features} &  {\footnotesize \hspace{-5mm} Avg. Acc.} \\ 
            \thickhline
            Audio-supervised CNN & 30.4\% \\
            ImageNet CNN & 42.0\% \\ 
            Sound & 45.8\% \\
            ImageNet + sound & 48.2\% \\
            ImageNet crop & 52.9\% \\
            Crop + sound & 59.4\% \\
            \thickhline
    \end{tabularx}}\\
\end{tabular}}
\vspace{-2mm}
  \caption{{\small (a) We ran variations of the full system and an
      image-matching method (RGB + spacetime). For each model, we
      include an oracle model (labeled with ``+ mat'') that draws its
      sound examples from videos with the same material label. (b)
      Class-averaged material recognition accuracy obtained by
      training an SVM with different image and sound features.}}
  \vspace{-6mm}
  \label{table:oracle-results}}
\end{figure}

\begin{figure}
  \vspace{-1mm}
  \includegraphics[width=\linewidth]{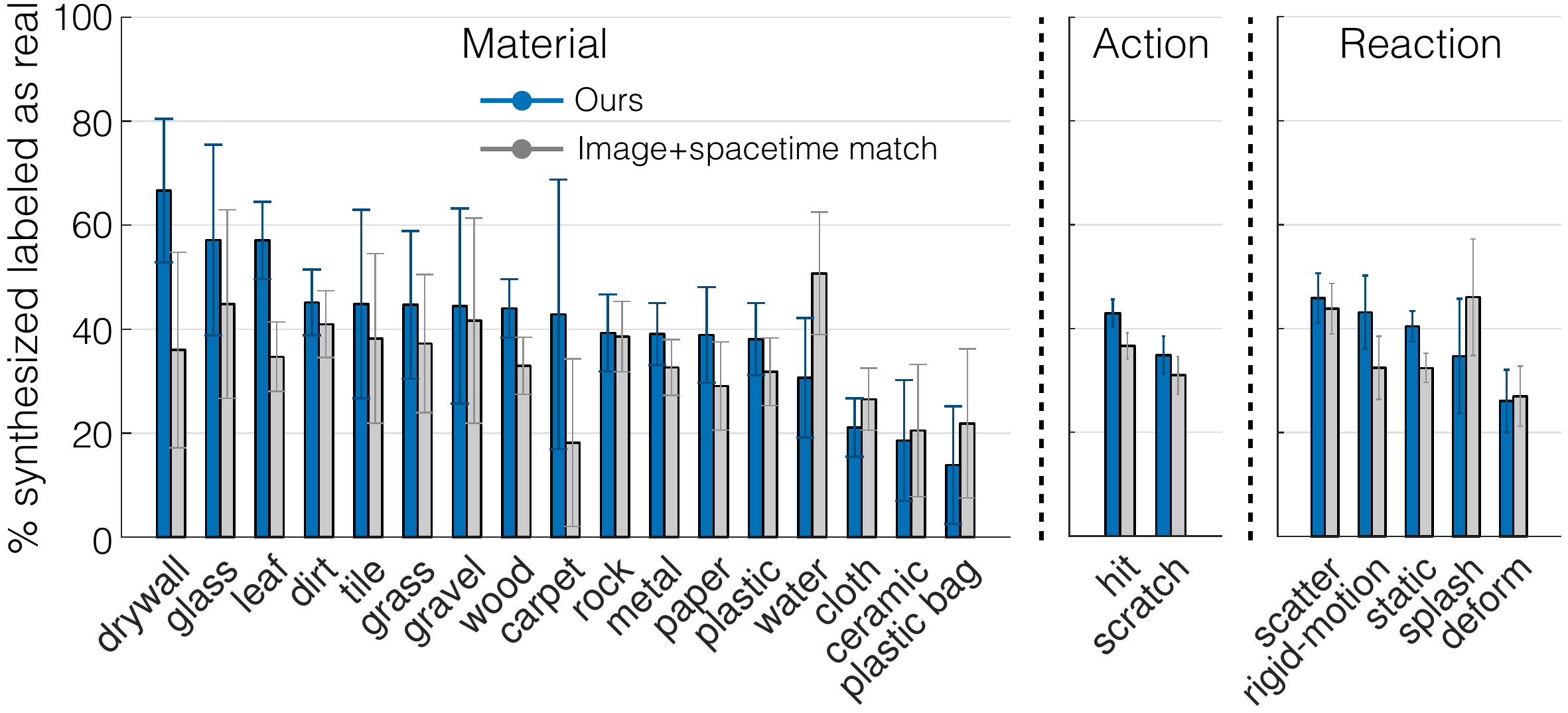}
\vspace{-7mm}
\caption{{\small Semantic analysis of psychophysical study. We show
    the rate that our algorithm fooled human participants for each
    material, action, and reaction class. The error bars show 95\%
    confidence intervals. Our approach significantly outperforms the
    highest-performing image-matching method (RGB + spacetime).}}
\label{fig:matacc}
\vspace{-4mm}

\end{figure}

We show results broken down by semantic category in \fig{fig:matacc}.
For some categories, such as grass and leaf, participants were
frequently fooled by our results.  Often when a participant was
fooled, it was because the sound prediction was simple and
prototypical (\eg, a simple thud noise), while the actual sound was
complex and atypical. True leaf sounds, for example, are highly varied
and may not be fully predictable from a silent video. When they are
struck, we hear a combination of the leaves themselves, along with
rocks, dirt, and whatever else is underneath them. In contrast, the
sounds predicted by our model tend to be closer to prototypical
grass/dirt/leaf noises.  Participants also sometimes made mistakes
when the onset detection failed, or when multiple impacts overlapped,
since this may have defied their expectation of hearing a single
impact.

We found that the model in which the RNN connections were broken was
often unable to detect the timing of the hit, and that it
under-predicted the amplitude of the sounds.  As a result, it
performed poorly on automated metrics and failed to find good matches.
The performance of our model with parametric waveform generation
varied widely between categories. It did well on materials such as
{\em leaf} and {\em dirt} that are suited to the relatively noisy
sounds that the method produces but poorly on hard materials such as
{\em wood} and {\em metal} (\eg, a confusion rate of \invdirt for dirt
and \invmetal for metal).  On the other hand, the example-based
approach was not effective at matching textural sounds, such as those
produced by splashing water (Fig. \ref{fig:matacc}).


\xpar{Auditory metrics}

We measured quantitative properties of sounds for the prediction
task. We chose metrics that were not sensitive to precise
timing. First, we measured the loudness of the sound, which we took to
be the maximum energy ($L_2$ norm) of the compressed subband envelopes
over all timesteps.  Second, we compared the sounds' spectral
centroids, which we measured by taking the center of mass of the
frequency channels for a one-frame (approx. 0.03 sec.) window around
the center of the impact.  We found that on both metrics, the network
was more accurate than the image-matching methods, both in terms of
mean squared error and correlation coefficients (\fig{fig:study}(a)).

\xpar{Oracle results} How helpful is material category information?  We conducted a second
study that controlled for material-recognition accuracy. Using the
subset of the data with material annotations, we created a model that
chose a random sound from the same class as the input video.  We also
created a number of oracle models that used these material labels
(\tbl{table:oracle-results}(a)).  For the best-performing
image-matching model (RGB + spacetime), we restricted the pool of
matches to be those with the same label as the input (and similarly
for the example-based synthesis method).  We also considered a model
that matched the ground-truth sound to the training set and
transferred the best match. We found that, while knowing the material
was helpful for each method, it was not sufficient, as the oracle
models did not outperform our model.  In particular, our model
significantly outperformed the random-sampling oracle ($p < 10^{-4}$).

\begin{figure*}[t]

    \vspace{-3mm}

    {\scriptsize \hspace{10mm} Frame from input video \hspace{14mm} Real vs. synthesized cochleagram\hspace{15mm}Frame from input video\hspace{16.5mm}Real vs. synthesized cochleagram} \vspace{1mm} \\
        {\centering \includegraphics[width=\hsize]{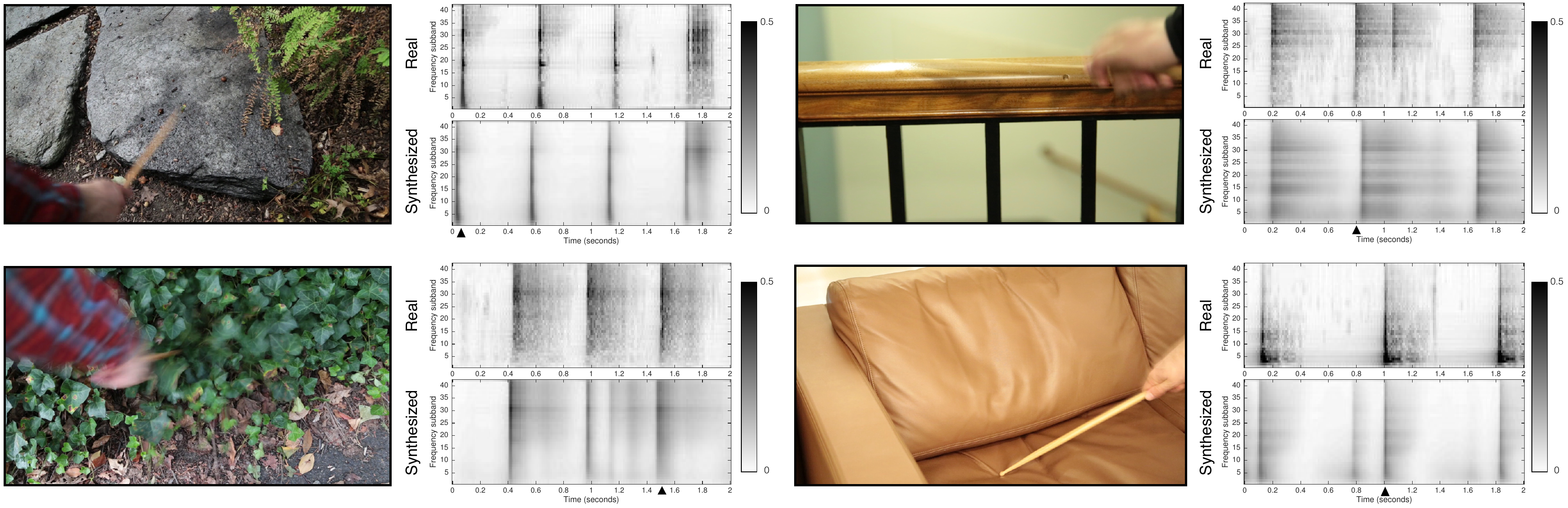}}
    
            \vspace{-3.25mm}
  \caption{{\small Automatic sound prediction results. We show
      cochleagrams for a representative selection of video sequences,
      with a sample frame from each sequence on the left. The frame is
      sampled from the location indicated by the black triangle on the
      $x$-axis of each cochleagram. Notice that the algorithm's
      synthesized cochleagrams match the general structure of the
      ground truth cochleagrams. Dark lines in the cochleagrams
      indicate hits, which the algorithm often detects. The algorithm
      captures aspects of both the temporal and spectral structure of
      sounds. It correctly predicts staccato taps in rock example and
      longer waveforms for rustling ivy. Furthermore, it tends to
      predict lower pitched thuds for a soft couch and higher pitched
      clicks when the drumstick hits a hard wooden railing (although
      the spectral differences may appear small in these
      visualizations, we evaluate this with objective metrics in
      Section \ref{experiments_section}). A common failure mode is
      that the algorithm misses a hit (railing example) or
      hallucinates false hits (cushion example). This frequently
      happens when the drumstick moves erratically. {\em Please see
        our video for qualitative results.}}}
  \vspace{-4mm}
 \label{fig:results}
\end{figure*}

\xpar{Impact detection} We also used our methods to produce sounds for
long, uncentered videos, a problem setting that allows us to evaluate
their ability to detect impact events.  We provide qualitative
examples in \fig{fig:results} and on our
webpage (\href{http://vis.csail.mit.edu}{vis.csail.mit.edu}). To
quantitatively evaluate its detection accuracy, we used the parametric
synthesis method to produce a waveform, applied a large gain to that
waveform, and then detected amplitude peaks (\sect{sec:dataset}). We
then compared the timing of these peaks to those of the ground truth,
considering an impact to be detected if a predicted spike occurred
within 0.1 seconds of it.  Using the predicted amplitude as a measure
of confidence, we computed average precision.  We compared our model
to an RGB-only model, finding that the spacetime images significantly
improve the result, with APs of \spacetimeap and \rgbap respectively.
Both models were pretrained with ImageNet.

\subsection{Learning about material and action by\\predicting sounds} 

By learning to predict sounds, did the network also learn something
about material and physical interactions?  To assess this, we tested
whether the network's output sounds were informative about material
and action class. We applied the same SVM that was trained to predict
material/action class on {\em real} sound features
(Sec. \ref{sec:subband}) to the sounds predicted by the model.  Under
this evaluation regime, it is not enough for the network's sounds to
merely be distinguishable by class: they must be close enough to real
sounds so as to be classified correctly by an SVM that has never seen
a predicted sound. To avoid the influence of pretraining, we used a
network that was trained from scratch. We note that this evaluation
method is different from that of recent unsupervised learning
models \cite{doersch2015unsupervised,agrawal2015learning,wang2015unsupervised}
that train a classifier on the network's feature activations, rather
than on a ground-truth version of the output.

Using this idea, we classified the material category from predicted
sound features. The classifier had class-averaged accuracy
of \predsoundbalacc, and its confusion matrix is shown in
Fig. \ref{fig:study}(b).
This accuracy indicates that our model learned an output
representation that was informative about material, even though it was
only trained to predict sound.  We applied a similar methodology to
classify action categories from predicted sounds,
obtaining \predhitscratchacc class-averaged accuracy (chance =
$50\%$), and \predreactionacc for classifying reaction categories
(chance = $20\%$).  We found that material and reaction recognition
accuracy improved with ImageNet pretraining (to \pretrainsoundbalacc
and to \pretrainreactionacc, respectively), but that there was a slight
decrease for action classification (to \pretrainhitscratchacc).



We also tested whether the predicted sound features convey information
about the hardness of a surface.  We grouped the material classes into
superordinate {\em hard} and {\em soft} classes, and trained a
classifier on real sound features (sampling 1300 examples per class),
finding that it obtained \predhardacc class-averaged accuracy (chance
= $50\%$). Here we have defined soft materials to be \{\emph{leaf,
grass, cloth, plastic bag, carpet}\} and hard materials to
be \{\emph{gravel, rock, tile, wood, ceramic, plastic, drywall, glass,
metal}\}.

We also considered the problem of directly predicting material class
from visual features. In \tbl{table:oracle-results}(b), we trained a
classifier using \fcseven features -- both those of the model trained
from scratch, and of a model trained on
ImageNet \cite{krizhevsky2012imagenet}. We concatenated color and
spacetime image features, since we found that this improved
performance.  We also considered an oracle model that cropped a
high-resolution ($256 \times 256$) patch from the impact location
using human annotations, and concatenated its features with those of
the full image (we used color images).  To avoid occlusions from the
arm or drumstick, we cropped the patch from the final frame of the
video.  We found that performing these crops significantly increased
the accuracy, suggesting that localizing the impact is important for
classification.  We also tried concatenating vision and sound features
(similar to \cite{Arnab_2015}), finding that this significantly
improved the accuracy.

The kinds of mistakes that the visual classifier (video $\rightarrow$
material) made were often different from those of the sound classifier
(sound $\rightarrow$ material). For instance, the visual classifier
was able to distinguish classes that have a very different appearance,
such as {\em paper} and {\em cloth}. These classes both make
low-pitched sounds (\eg, cardboard and cushions), and were sometimes
are confused by the sound classifier. On the other hand, the visual
classifier was more likely to confuse materials from outdoor scenes,
such as rocks and leaves -- materials that sound very different but
which frequently co-occur in a scene. When we analyzed our model by
classifying its sound predictions (video $\rightarrow$ sound
$\rightarrow$ material), the resulting confusion matrix
(Fig. \ref{fig:study}(b)) contains both kinds of error: there are {\em
visual analysis} errors when it misidentifies the material that was
struck, and {\em sound synthesis} errors when it produces a sound that
was not a convincing replica of the real sound.

\vspace{-1.5mm}
\section{Discussion}

In this work, we proposed the problem of synthesizing visually
indicated sounds -- a problem that requires an algorithm to learn
about material properties and physical interactions. We introduced a
dataset for studying this task, which contains videos of a person
probing materials in the world with a drumstick, and an algorithm
based on recurrent neural networks.  We evaluated the quality of our
approach with psychophysical experiments and automated metrics,
showing that the performance of our algorithm was significantly better
than baselines. 

We see our work as opening two possible directions for future
research. The first is producing realistic sounds from videos,
treating sound production as an end in itself.  The second direction
is to use sound and material interactions as steps toward physical
scene understanding.

\xpar{Acknowledgments.} {\small This work was supported by NSF grants 6924450 and 6926677,
by Shell, and by a Microsoft Ph.D. Fellowship to A.O.  We thank Carl
Vondrick and Rui Li for the helpful discussions, and the workers at
Middlesex Fells, Arnold Arboretum, and Mt. Auburn Cemetery for not
asking too many questions while we were collecting the {\em Greatest
Hits} dataset.}

{\footnotesize \bibliographystyle{ieee} \bibliography{vis} }

\renewcommand{\thesection}{A\arabic{section}}
\renewcommand{\thefigure}{A\arabic{figure}}
\renewcommand{\theHsection}{othersection\thesection}
\setcounter{figure}{0}
\setcounter{section}{0}





\section{Model implementation}

\label{sec:methoddetails}

We provide more details about our model and sound representation.


\subsection{Detection model}

\label{sec:detmodel}

We describe the variation of our model that performs the detection
task (\sect{sec:predtasks}) in more detail.

\xpar{Timing}

We allow the sound features to undergo small time shifts in order to
account for misalignments for the detection
task. During each iteration of backpropagation, we shift the sequence
so as to minimize the loss in Equation \ref{eq:loss}. We resample the
feature predictions to create a new sequence $\hat{\vec{s}}_1,
\hat{\vec{s}}_2,...,\hat{\vec{s}}_T$ such that $\hat{\vec{s}}_t =
\sh_{t + L_t}$ for some small shift $L_t$ (we use a maximum shift of 8
samples, approximately 0.09 seconds). During each iteration, we infer
this shift by finding the optimal labeling of a Hidden Markov Model:
\vspace{-2.5mm}
\begin{equation}
\sum\limits_{t = 1}^T w_t \rho(\norm{\hat{\vec{s}}_t - \sg_t}) + V(L_t, L_{t+1}),
\vspace{-2mm}
\end{equation}
where $V$ is a smoothness term for neighboring shifts. For this, we
use a Potts model weighted by $\frac{1}{2}(\norm{\sg_t} +
\norm{\sg_{t+1}})$ to discourage the model from shifting the sound
near high-amplitude regions.  We also include a weight variable $w_t =
1 + \alpha \delta(\tau \leq ||\sg_t||)$ to decrease the importance of
silent portions of the video (we use $\alpha = 3$ and $\tau =
2.2$). During each iteration of backpropagation, we align the two
sequences, then propagate the gradients of the loss to the shifted
sequence.

To give the RNN more temporal context for its predictions, we also
delay its predictions, so that at frame $f$, it predicts the sound
features for frame $f-2$.

\xpar{Transforming features for neighbor search}

For the detection task, the statistics of the synthesized sound
features can differ significantly from those of the ground truth --
for example, we found the amplitude of peaks in the predicted
waveforms to be smaller than those of real sounds. We correct for
these differences during example-based synthesis (\sect{sec:waveform})
by applying a coloring transformation before the nearest-neighbor
search. More specifically, we obtain a whitening transformation for
the predicted sound features by running the neural network on the test
videos and estimating the empirical mean and covariance at the
detected amplitude peaks, discarding peaks whose amplitude is below a
threshold. We then estimate a similar transformation for ground-truth
amplitude peaks in the training set, and we use these transformations
to color (\ie transform the mean and covariance of) the predicted
features into the space of real features before computing their $L_1$
nearest neighbors. To avoid the influence of multiple, overlapping
impacts on the nearest neighbor search, we use a search window that
starts at the beginning fo the amplitude spike.



\xpar{Evaluating the RNN for long videos} When evaluating our model on
long videos, we run the RNN on 10-second subsequences that overlap by
30\%, transitioning between consecutive predictions at the time that
has the least sum-of-squares difference between the overlapping
predictions.

\subsection{Sound representation}

\label{sec:appsoundrep}

\begin{figure}[t]
  {\centering
    {\includegraphics[width=\linewidth]{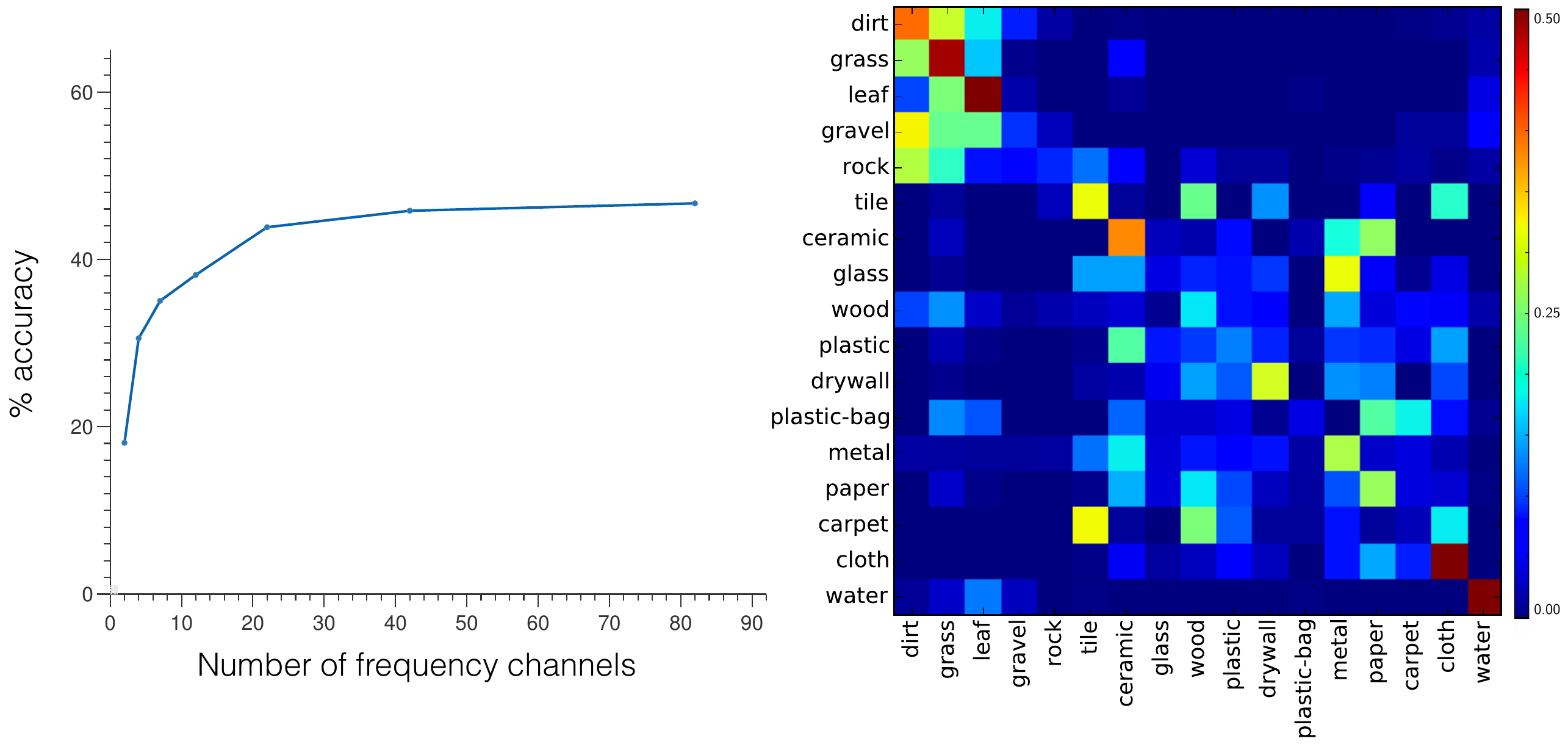}} 
    {\footnotesize \phantom{~~~~~}(a) Sound recognition accuracy ~~~~~~~~~(b) Predicted sound confusions}}
\caption{\small (a) Class-averaged accuracy for recognizing materials,
  with an SVM trained on real sounds. We varied the number of
  band-pass filters and adjusted their frequency spacing accordingly
  (we did not vary the temporal sampling rate). (b) Confusion matrix
  obtained by classifying the sounds predicted by our pretrained
  model, using a classifier trained on real sound features (\cf the
  same model without pretraining in \fig{fig:study}(b).)}

\vspace{-3mm}
\label{fig:channel}
\end{figure}

We measured performance on the task of assigning material labels to
ground-truth sounds after varying the number frequency channels in the
subband envelope representation.  The result is shown in
\fig{fig:channel}.  To obtain the ordering of material classes used in
visualizations of the confusion matrices (\fig{fig:mean-coc}), we
iteratively chose the material category that was most similar to the
previously chosen class.  When measuring the similarity between two
classes, we computed Euclidean distance between rows of a (soft)
confusion matrix -- one whose rows correspond to the mean probability
assigned by the classifier to each target class (averaged over all
test examples).



\subsection{Network structure}

\begin{figure*}[t]
  \centering
  \includegraphics[width=\linewidth]{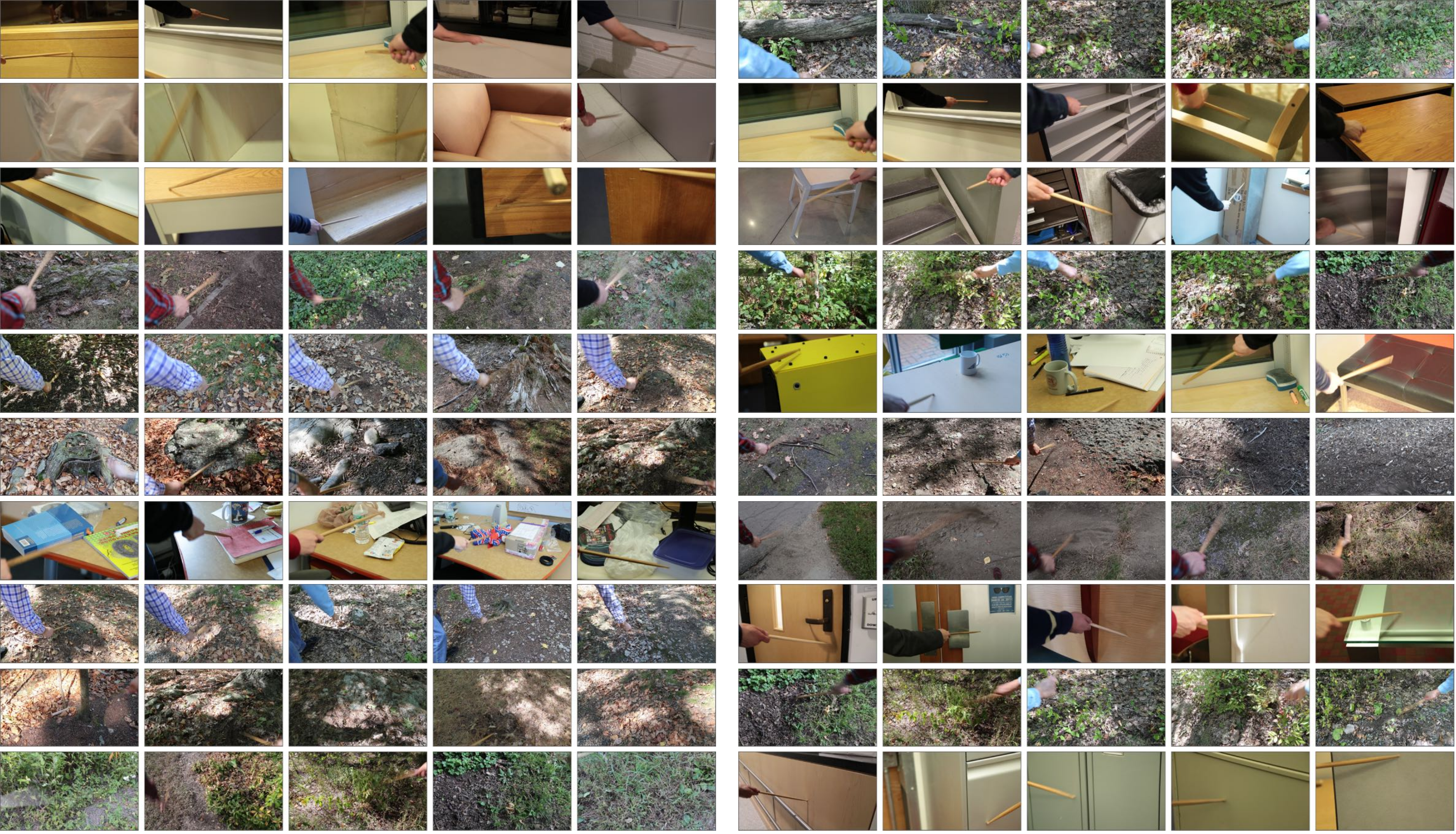}
\caption{\small A ``walk'' through the dataset using AlexNet \fcseven
    nearest-neighbor matches.  Starting from the left, we matched an
    image with the database and placed its best match to its right. We
    repeat this 5 times, with 20 random initializations. We used only
    images taken at a contact point (the middle frames from the
    ``centered'' videos). To avoid loops, we removed videos when any
    of their images were matched.  The location of the hit, material,
    and action often vary during the walk. In some sequences, the arm
    is the dominant feature that is matched between scenes.}
\label{fig:walk}
\end{figure*}

We used AlexNet \cite{krizhevsky2012imagenet} for our CNN
architecture.  For the pretrained models, we precomputed the {\em
  pool5} features and fine-tuned the model's two fully-connected
layers.  For the model that was trained from scratch, we applied batch
normalization \cite{ioffe2015batch} to each training mini-batch. For
the centered videos, we used two LSTM layers with a 256-dimensional
hidden state (and three for the detection model). When
using multiple LSTM layers, we compensate for the difference in video
and audio sampling rates by upsampling the input to the last LSTM
layer (rather than upsampling the CNN features), replicating each
input $k$ times (where again $k = 3$).

\section{Dataset details}

\label{sec:datasetdetails}
  
In \fig{fig:walk}, we show a ``walk'' through the dataset using {\em
  fc7} features, similar to \cite{zitnickvideo}.  Our data was
collected using two wooden (hickory) drumsticks, and an SLR camera with a
29.97 Hz framerate.  We used a ZOOM H1 external audio recorder, and a
Rode VideoMic Pro microphone. Online workers labeled the impacts by
visually examining silent videos, without sound.  We gave them
finer-grained categories than, then merged similar categories that
were frequently labeled inconsistently by workers.  Specifically, we
merged {\em cardboard} and {\em paper}; {\em concrete} and {\em rock};
{\em cloth} and {\em cushion} (often the former physically covers the
latter); and {\em rubber} and {\em plastic}. To measure overall
consistency between workers, we labeled a subset of the impacts with 3
or more workers, finding that their material labels agreed with the
majority \humanconsistency of the time on the fine-grained categories.
Common inconsistencies include confusing {\em dirt} with {\em leaf}
(confused 5\% of the time); {\em grass} with {\em dirt} and {\em leaf}
(8\% each); and {\em cloth} with (the fine-grained category) {\em
  cushion} (9\% of the time).






\end{document}